\documentclass[runningheads]{llncs}
\usepackage[T1]{fontenc}

\usepackage{graphicx}
\usepackage{algorithm}
\usepackage{algorithmic}
\usepackage{subfig}
\usepackage{glossaries}
\usepackage{amsmath,amssymb,amsfonts}
\usepackage{comment}
\usepackage{wrapfig}

\newacronym{ems}{EMS}{Energy Management System}
\newacronym{esu}{ESU}{Energy Storage Units}
\newacronym{soc}{SoC}{State of Charge}
\newacronym{ppo}{PPO}{Proximal Policy Optimization}
\newacronym{rl}{RL}{Reinforcement Learning}
\newacronym{nsl}{NSL}{Non Shiftable Load}
\newacronym{fe}{FE}{Feature Extractor}

\begin{document}
%
%\title{Meta-Learning with Inter-Task Knowledge Transfer for
%Energy Management Systems}

\title{Meta-RL with Shared Representations Enables Fast Adaptation in Energy Systems}

\titlerunning{Meta-RL with Shared Representations}

\author{Th\'eo Zangato\inst{1}\thanks{Corresponding author. Email: zangato@lipn.univ-paris13.fr} \and
Aomar Osmani\inst{2} \and
Pegah Alizadeh\inst{1,3}}

\authorrunning{T. Zangato et al.}

\institute{
Universit\'e Sorbonne Paris Nord, France \and
INSA Rouen, France \and
Université Lumière Lyon 2, Université Claude Bernard Lyon 1, ERIC, 69007, Lyon, France
}
%\and
\maketitle              % typeset the header of the contribution
\begin{abstract}
Meta-Reinforcement Learning addresses the critical limitations of conventional Reinforcement Learning in multi-task and non-stationary environments by enabling fast policy adaptation and improved generalization. We introduce a novel Meta-RL framework that integrates a bi-level optimization scheme with a hybrid actor-critic architecture specially designed to enhance sample efficiency and inter-task adaptability. To improve knowledge transfer, we meta-learn a shared state feature extractor jointly optimized across actor and critic networks, providing efficient representation learning and limiting overfitting to individual tasks or dominant profiles. Additionally, we propose a parameter-sharing mechanism between the outer- and inner-loop actor networks, to reduce redundant learning and accelerate adaptation during task revisitation. 
The approach is validated on a real-world Building Energy Management Systems dataset covering nearly a decade of temporal and structural variability, for which we propose a task preparation method to promote generalization. Experiments demonstrate effective task adaptation and better performance compared to conventional RL and Meta-RL methods. %152

\keywords{Reinforcement Learning  \and Meta Learning \and Energy Management Systems}
\end{abstract}
\section{Introduction}

\acrlong{ems}s (EMS) are central to improving energy efficiency, reducing operational costs, and achieving sustainability goals. As energy systems become increasingly complex, integrating distributed renewables, variable demands, and real-time control, intelligent \acrshort{ems} solutions are essential for adaptive, scalable, and data-driven decision-making across buildings and grids. \acrfull{rl} has shown strong potential for EMS control by learning flexible policies under uncertainty~\cite{chen2022reinforcement}. However, conventional \acrshort{rl} methods often struggle to generalize across heterogeneous building environments and temporal variations (e.g., seasonal or occupancy patterns) and require extensive interactions, making them impractical for real-world deployments with costly feedback.

Meta-Reinforcement Learning (Meta-RL) addresses these limitations by enabling agents to transfer prior knowledge and rapidly adapt to new tasks drawn from a shared distribution~\cite{finn2017model,nichol2018first}. Gradient-based approaches such as MAML~\cite{finn2017model} and Reptile~\cite{nichol2018first} pioneered this direction, while CAVIA~\cite{zintgraf2019fast} introduced context variables to separate shared and task-specific parameters.
Subsequent work has explored parameter sharing and meta-gradient strategies. Meta-Gradient RL~\cite{xu2018meta} and Metatrace~\cite{young2018metatrace} adapt hyperparameters online, FLAP~\cite{peng2021linear} employs shared linear policies with lightweight adapters and MACAW~\cite{mitchell2021offline} stabilizes offline learning via advantage-weighted regression. PEARL~\cite{rakelly2019efficient} and RL$^2$~\cite{duan2016rl} improve task inference through latent contexts or recurrent policies. However, these methods underexploit shared actor–critic structures across tasks, limiting their ability to leverage inter-task regularities in structured domains.

Meta-RL has recently been applied to EMS control. \cite{kurte2020evaluating} showed improved adaptation for HVAC systems, \cite{shen2024real} combined MAML with SAC for microgrid scheduling, and MetaEMS~\cite{zhang2022metaems} aggregated experiences across buildings for joint updates. \cite{xiong2023meta} demonstrated that meta-trained scheduling policies can generalizing to unseen loads and renewables.%, while \cite{lu2024meta} proposed a Reptile-based multi-objective framework for appliance scheduling under renewable uncertainty. 
While MAML-based approaches dominate, they require costly full-model gradient updates. First-order alternatives such as Reptile are more scalable but often fail to retain task-specific knowledge, focusing on task differentiation rather than knowledge consolidation. In EMS environments, where tasks exhibit high structural similarity and low conflict, maximizing shared information is crucial to improve generalization and reduce sample complexity. We focus on two underexplored EMS Meta-RL challenges:
\begin{itemize}
    \item \textbf{Task selection} for robust generalization. In EMS, tasks vary temporally (e.g., day–night cycles) and spatially (eg., across zones). Identifying diverse but representative tasks is crucial to define the learning scope and accelerate adaptation~\cite{jose2021information}. %We propose a hybrid task selection method to capture periodic and workload variability in energy consumption patterns.
    \item \textbf{Architectures for knowledge sharing and policy reuse.} Unlike typical Meta-RL benchmarks involving heterogeneous tasks~\cite{kirsch2019improving}, EMS tasks share structural consistency and exhibit low inter-task conflict. %Leveraging this property, we design a hybrid actor–critic Meta-RL framework that captures shared dynamics across tasks.
\end{itemize}

We propose: 1) a meta-learned shared feature extractor enabling transferable representation learning, 2) a task-specific actor reuse mechanism that reduces redundant exploration and improves sample efficiency, 3) a task selection strategy that balances diversity and similarity.

\section{Proposed approach}

\subsection{Preliminaries}
\label{sec:preliminaries}
Reinforcement Learning (RL) operates in environments modeled as a Markov Decision Process (MDP) $\mathcal{M} = \langle S,A,P,R,\gamma \rangle$, where $S$ and $A$ are the state and action spaces, $P(s'|s,a)$ the transition dynamics, $R(s,a)$ the reward function, and $\gamma \in [0,1]$ the discount factor. The goal is to learn a policy $\pi: S \rightarrow A$ maximizing the expected discounted return:
%\begin{equation} \label{eq:rl-goal}
$\mathcal{J} = \mathbb{E}_{\tau \sim \pi} \left[ \sum_{t=0}^{T} \gamma^t r_t \right]$,
%\end{equation}
where $\tau = {(s_t,a_t,r_t)}_{t=0}^{T}$ denotes a trajectory.
In real-world scenarios, agents often need to quickly adapt across related tasks. Meta-Reinforcement Learning (Meta-RL) addresses this by sampling $N$ tasks $\{\mathcal{M}_i\}_{i=1}^N$ from a distribution $p(\mathcal{M})$, where each task defines its own MDP. For each task, a dataset $D_i \in ((S_i \times A_i \times \mathbb{R})^T)^H$ of $H$ episodes is collected, and a task-specific learner $f_i(D_i)$ produces a policy $\pi(a|s)$ parameterized by $\Theta_i$. The meta-learner $\mathcal{F}$ is trained to generate such learners,
$\mathcal{F}(\{D_i\}_{i=1}^N): \{((S_i \times A_i \times \mathbb{R})^T)^H\}_{i=1}^N \to \{f_i\}_{i=1}^N$,
enabling fast adaptation to unseen tasks $\mathcal{M}_{\text{new}} \sim p(\mathcal{M})$ via $f_{\text{new}} = \mathcal{F}(D_{\text{new}})$ \cite{beck2023survey}.

\subsection{Critic Feature Extractor Meta Learning (CFE)}
\label{section:critic-feautre-extractor}

\begin{figure}
    \centering
    \includegraphics[scale=0.5]{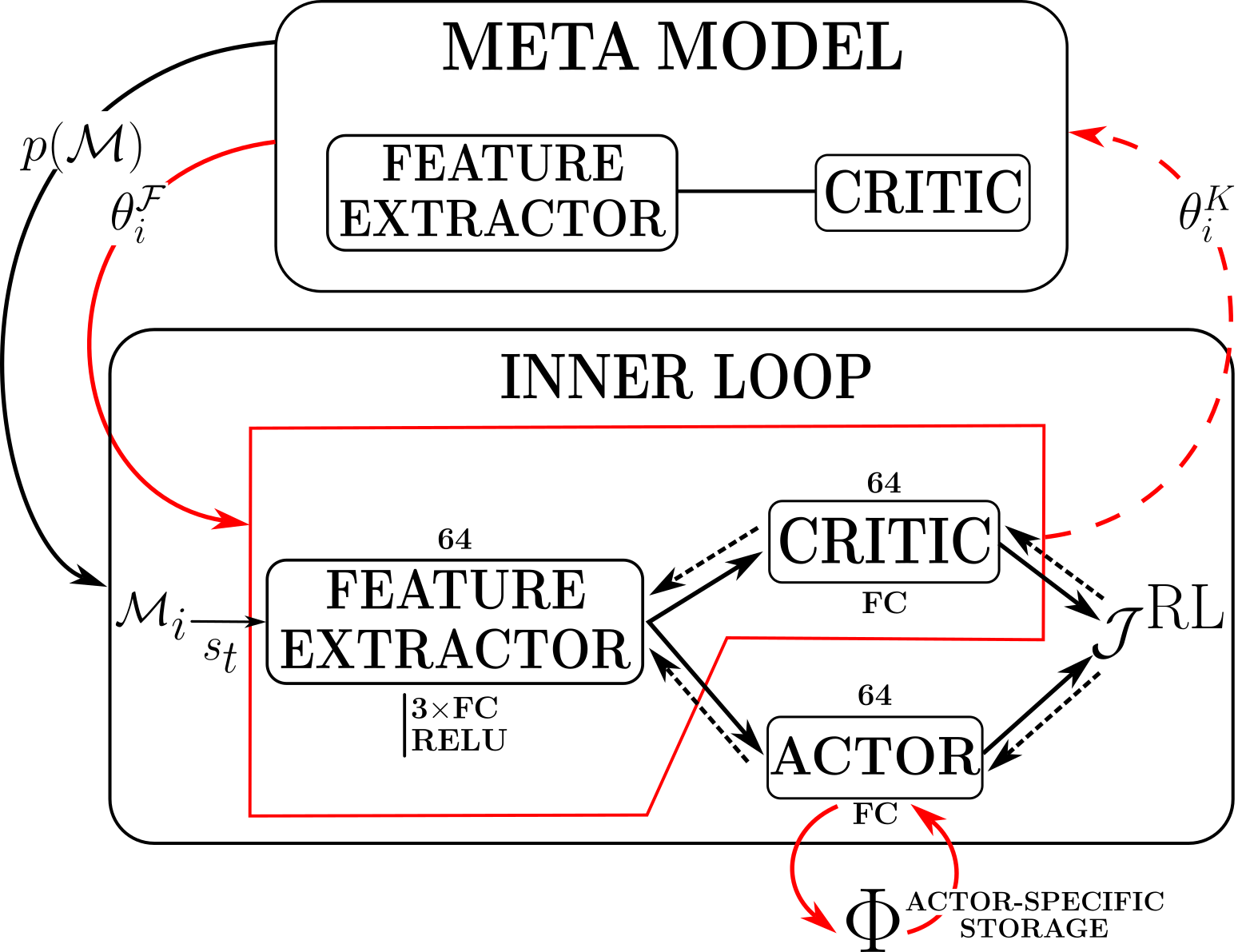}
    \caption{Model architecture. Red arrows show interactions between the inner and outer loops where each task knowledge is propagated to the meta-model, dashed arrows indicate gradient flow.}
    \label{fig:archi}
\end{figure}

% \begin{wrapfigure}{r}{0.5\textwidth}
%   \begin{center}
%     \includegraphics[scale=0.5]{figs/meta-archi3.png}
%   \end{center}
%   \caption{Model architecture. Red arrows show interactions between the inner and outer loops where each task knowledge is propagated to the meta-model, dashed arrows indicate gradient flow.}
%     \label{fig:archi}
% \end{wrapfigure}

We propose a Meta-RL framework (CFE) with task selection to optimize control policies for \acrshort{ems}.
Unlike typical Meta-RL settings where tasks differ structurally (e.g., navigation vs. manipulation), our tasks are instances of the same control problem. Variability comes from exogenous factors (e.g., temperature, irradiation), temporal patterns (e.g., seasons), and operational contexts, all observable by the agent.
Because tasks share similar dynamics and exhibit no inter-task conflict, our objective is to capture shared representations while enabling rapid adaptation to new conditions.

Our method includes two key components. First, we meta-learn a shared feature encoder across actor–critic networks. The encoder extracts latent representations that generalize across tasks, improving critic value estimation and accelerating actor adaptation. Second, we introduce an actor reuse mechanism that stores task-specific actor parameters and reuses them when the same task reoccurs, reducing redundant exploration and improving sample efficiency. The overall architecture is depicted in Figure~\ref{fig:archi}. The code is available on github\footnote{\url{https://github.com/zxnga/meta-rl-ems}}.

%Given the high structural similarity among EMS tasks, we exploit shared representations to facilitate cross-task generalization. The low inter-task conflict allows for the construction of a common feature space that captures invariant environmental dynamics.
%Building upon the concept of shared model layers \cite{andrychowicz2021matters,cobbe2021phasic}, we extend it to our meta-learning architecture by introducing a shared feature extractor, enabling representation-level transfer without requiring full parameter propagation to the meta-learner. The overall architecture is depicted in Figure~\ref{fig:archi}.%, and the training procedures are detailed in Algorithms~\ref{alg:meta-train} and~\ref{alg:meta-test}.

\subsection{Adaptation Stage}
The inner loop treats each task as a standalone \acrshort{rl} problem. Its objective is to train a task-specific agent that produces parameters adapted to the current task.
For a new task $\mathcal{M}_i \sim p(\mathcal{M})$, the learner $f_i$ is initialized with parameters inherited from the meta-learner $\mathcal{F}$ based on the sampled task set $\mathcal{S}=\{\mathcal{M}_j|j \text{ sampled}\}$ such as $\theta_i^\mathcal{F}=\mathcal{F}\left(\{\theta_j^{\mathcal{M}_i}\}_{\mathcal{M}_{j\in\mathcal{S}}}\right)$. This initialization holds knowledge from the previously sampled tasks, thus $f_i$ is not trained from scratch \cite{finn2017model}.

%\paragraph{Actor-Critic Shared Representation}
Given the high structural similarity of \acrshort{ems} tasks, we leverage shared representations to promote cross-task generalization. The low inter-task conflict enables learning a common feature space that captures invariant environment dynamics. Inspired by shared model layers \cite{andrychowicz2021matters,cobbe2021phasic}, we extend this concept to our meta-learning framework through a shared feature extractor, allowing representation-level transfer without full parameter propagation to the meta-learner. %Actor and critic networks share a common \acrfull{fe} that encodes a joint state representation, retaining actor knowledge in shared layers. This avoids  the need to propagate actor weights to the meta-model while still improves inner-loop adaptability through knowledge transfer. 

Each base learner $f_i=\{\psi_{\theta_i^\psi}, \pi_{\theta_i^\pi}, Q_{\theta_i^Q}\}$ is defined by actor and critic specific layers $\pi_{\theta_i^\pi}$ and $Q_{\theta_i^Q}$ respectively and a common \acrfull{fe} $\psi_{\theta_i^\psi}$. The later produces a learned representation $z=g^\psi(s)$ of the input state, used by both actor and critic networks. The learner $f_i$ updates its parameters using task-specific data $D_i$ and the objective function $J(\theta_i^\mathcal{F}, D_i)$ as a standard single-agent RL process that we define generically to be model agnostic:
\begin{equation}
\theta_i^{X, \text{new}} = \theta_i^{X, (t_\phi)} + \alpha \mathbb{E}_{\tau\sim\pi}\left[ \nabla_{\theta^X_i} \mathcal{J}^{X}(\theta_i^{X, (t_\phi)})\right]
\label{eq:general-update}
\end{equation}
where $X \in \{\pi, Q, \psi\}$ represents the actor, critic, or \acrshort{fe}, $\theta_i^{X, (t_\phi)}$ its current corresponding parameters for task $i$ at meta-training step $t_\phi$, and $\mathcal{J}^{X}$ its objective function.
After $K$ inner-loop updates, the adaptation stage $f_i(D_i;\theta_i^\mathcal{F})=\theta_i^{K}$ outputs a task-specific agent with parameters $\theta_i^K$.

\subsection{Meta-Training}

The outer loop aims to optimize the meta-learner $\mathcal{F}_\phi$, which provides improved initialization parameters $\theta_i^\mathcal{F}$ for the task-specific learners $f_i$, enabling faster and more efficient adaptation. Following prior work \cite{rakelly2019efficient,sung2017learning}, we choose to meta-optimize only a subset of model parameters. Specifically, only the parameters corresponding to the shared feature extractor $\phi^\psi$ and the critic-specific layers $\phi^Q$ are propagated to the meta-learner $\mathcal{F}=\{\psi_{\phi^\psi}, Q_{\phi^Q} \}$. 
This is motivated based on the asymmetry between the actor and critic roles in \acrshort{rl}. The critic plays a supporting role by estimating expected returns from stable signals across tasks, while the actor directly affects the specific policy needed to solve the task. The critic benefits more meta-learning given its generic role across all tasks.

The meta-learner $\mathcal{F}=\{\psi_{\phi^\psi}, Q_{\phi^Q} \}$ maintains and updates the meta-weights %and optimizes the meta-objective $\mathcal{J}_{meta}(\phi^\psi, \phi^Q) = \mathbb{E}_{\mathcal{M} \sim p(\mathcal{M})} \left[\mathbb{E}_{D^\mathcal{M}} \left[\mathcal{J}(\theta^\mathcal{M}) \right] \right]$ 
allowing the model to generalize across the distribution $p(\mathcal{M})$. For each meta-iteration we sample a batch of $M$ tasks $\mathcal{B}=\{\mathcal{M}_i\}^M_{i=1}$ from $p(\mathcal{M})$. Once the updated parameters $\theta_i^K$ are obtained for each of the $M$ tasks, we propagate the parameters to the meta-model. The meta-learner $\mathcal{F}_\phi$ uses the Reptile update rule, which approximates the first-order gradient of the meta-objective \cite{nichol2018first}:
\begin{equation}
    \mathcal{J}_{meta}(\phi^\psi, \phi^Q) = \mathbb{E}_{\mathcal{M} \sim p(\mathcal{M})} \left[\mathbb{E}_{D^\mathcal{M}} \left[\mathcal{J}(\theta^\mathcal{M}) \right] \right]
\end{equation}
%$\mathcal{J}_{\text{meta}}(\phi) = \mathbb{E}_{\mathcal{M} \sim p(\mathcal{M})} \allowbreak \left[\mathbb{E}_{D^\mathcal{M}} \left[\mathcal{J}(\theta^\mathcal{M}) \right] \right]$, 
% where $\mathcal{J}(\theta^{\mathcal{M}})$ is given in Equation~\ref{eq:rl-goal} \cite{nichol2018first}.

The meta-update is computed as the average difference between the adapted task-specific ($\theta_i^{K}$) and initialization ($\phi$) parameters, across all tasks in $\mathcal{B}$. The meta-learner parameters are updated as:
\begin{equation} \label{eq:reptile}
\Delta\phi = \frac{1}{M} \sum_{i=1}^M \left( \theta_i^{K} - \phi \right), \quad
\phi \leftarrow \phi + \alpha^{\text{outer}} \Delta\phi,
\end{equation}

%The update procedure of each sampled batch of tasks $\mathcal{B}$ is repeated for each meta-training step $t_\phi$, with the meta-parameters $\phi^{(t_\phi)}$ reflecting the accumulated knowledge from all tasks sampled in previous meta-training steps $(j<t_\phi)$ such as $\mathcal{S} = \{\mathcal{M}_j|j<t_\phi\}$:

At each meta-training step $t_\phi$, a new batch of tasks $\mathcal{B}^{(t_\phi)}$ is sampled, and the meta-parameters $\phi^{(t_\phi)}$ are updated. Let $\mathcal{S} = \{\mathcal{M}_j|j<t_\phi\}$ be the set of sampled task up to $t_\phi$, The meta-parameters $\phi^{(t_\phi)}$ reflect the accumulated knowledge from all tasks sampled in previous meta-training steps. Their update is expressed as:
\begin{equation}
    \phi^{(t_\phi)} = \mathcal{F}\left( \left\{\theta_i^{K,(j)} | i \in \mathcal{B}^{(j)}, j<t_\phi \right\}\right)
\end{equation}
%where $\{\theta_i^{K,(j)}\}$ are the task-specific parameters adapted during meta-training steps $j$ and $\mathcal{B}^{(j)}$ the batch of tasks at step $j$.

This ensures the meta-learner accumulates knowledge over time to provide better initialization for future tasks.

\subsection{Inner Loop Actor Weights Reuse (AR)}

While actor weights are not propagated to the meta-model $\mathcal{F}$, we store task-specific actor parameters across meta-training iterations to retain prior knowledge. This is particularly beneficial for tasks with long temporal dependencies, where effective policies (e.g., charging–discharging cycles) require extensive training. Reusing trained weights 1) improves sample efficiency by focusing exploration on critical states, 2) prevents relearning of common behaviors, and 3) aligns meta-learned parameters with previously optimized task policies. For each task $\mathcal{M}_i$, the adapted actor parameters $\theta_i^{\pi,K}$ are stored and reused when the same task reappears:
\begin{equation}
    \theta_i^{\pi,(t*\phi)} =
    \begin{cases}
        \text{sampled from } \mathcal{P}_\pi, & \text{if } \mathcal{M}_i \notin \mathcal{S}, \\
        \Phi^\pi(\mathcal{M}_i), & \text{if } \mathcal{M}_i \in \mathcal{S},
    \end{cases}
\end{equation}
where $\Phi^\pi: \mathcal{M} \to \mathbb{R}$ maps each task to its stored actor parameters.
After adaptation, $\mathcal{S}$ is updated as $\mathcal{S} \leftarrow \mathcal{S} \cup {\mathcal{M}_i}$ and $\Phi^\pi(\mathcal{M}_i) \gets \theta_i^{\pi,\text{new}}$.
This balances \textbf{specialization}, via task-specific reuse of $\Phi^\pi(\mathcal{M}_i)$ for known tasks, and \textbf{generalization}, via shared meta-parameters $(\phi^\psi, \phi^Q)$ that enable rapid adaptation to unseen tasks.

% \begin{algorithm}[ht]
% \caption{Meta-Testing Stage for Meta-RL}\label{alg:meta-test}
% \begin{algorithmic}[1]
% \REQUIRE Unseen task $\mathcal{M}_{\text{new}}\sim p(\mathcal{M})$, meta-learned parameters $\phi^\psi, \phi^Q$, random distribution $\mathcal{P}$, inner-loop updates $K$
% \STATE Initialize $f_{\text{new}}$: $\theta^{\psi} \gets \phi^\psi$, $\theta^Q \gets \phi^Q$, $\theta^\pi \sim \mathcal{P}$
% \FOR{$k = 1$ \textbf{to} $K$}
%     \STATE Collect data $D_{\text{new}}$ in $\mathcal{M}_{\text{new}}$ using policy $\pi(a|s;\theta_\pi, \theta_{\psi})$
%     \STATE Update $f_{\text{new}}$ using \ref{eq:general-update}
% \ENDFOR
% \STATE \textbf{Output:} Adapted task-specific learner $f_{\text{new}}$ for $\mathcal{M}_{\text{new}}$
% \end{algorithmic}
% \end{algorithm}

\subsection{Meta-RL Task Selection and Evaluation Protocol}

To promote task diversity, we cluster buildings to identify distinct consumption behavior profiles following the method in \cite{zangato2025data}. For each time series $\mathbf{x}_i \in \mathbb{R}^m$, we compute a smoothed derivative $\mathbf{x}'_i(t)$ via cubic spline interpolation and derive its frequency-domain signature as the magnitude of the Fourier transform: $\hat{\mathbf{x}}_i = |F[\mathbf{x}'_i(t)]|$. Each spectrum is normalized to unit norm, $\tilde{\mathbf{x}}_i = \frac{\hat{\mathbf{x}}_i}{\|\hat{\mathbf{x}}_i\|}$, and pairwise similarity is computed using cosine distance $D_{ij} = 1 - \frac{\hat{\mathbf{x}}_i \cdot \hat{\mathbf{x}}_j}{\|\hat{\mathbf{x}}_i\| \, \|\hat{\mathbf{x}}_j\|}$. %We apply \textbf{hierarchical agglomerative clustering} with average linkage on the cosine-distance matrix $\mathbf{D} \in \mathbb{R}^{n \times n}$.
The resulting matrix $\mathbf{D} \in \mathbb{R}^{n \times n}$ serves as input to \textbf{hierarchical agglomerative clustering} with average linkage $d_{\text{avg}}(A, B) = \frac{1}{|A| \cdot |B|} \sum_{x \in A} \sum_{y \in B} D(x, y)$.

\begin{figure}
    \centering
    \includegraphics[scale=0.4]{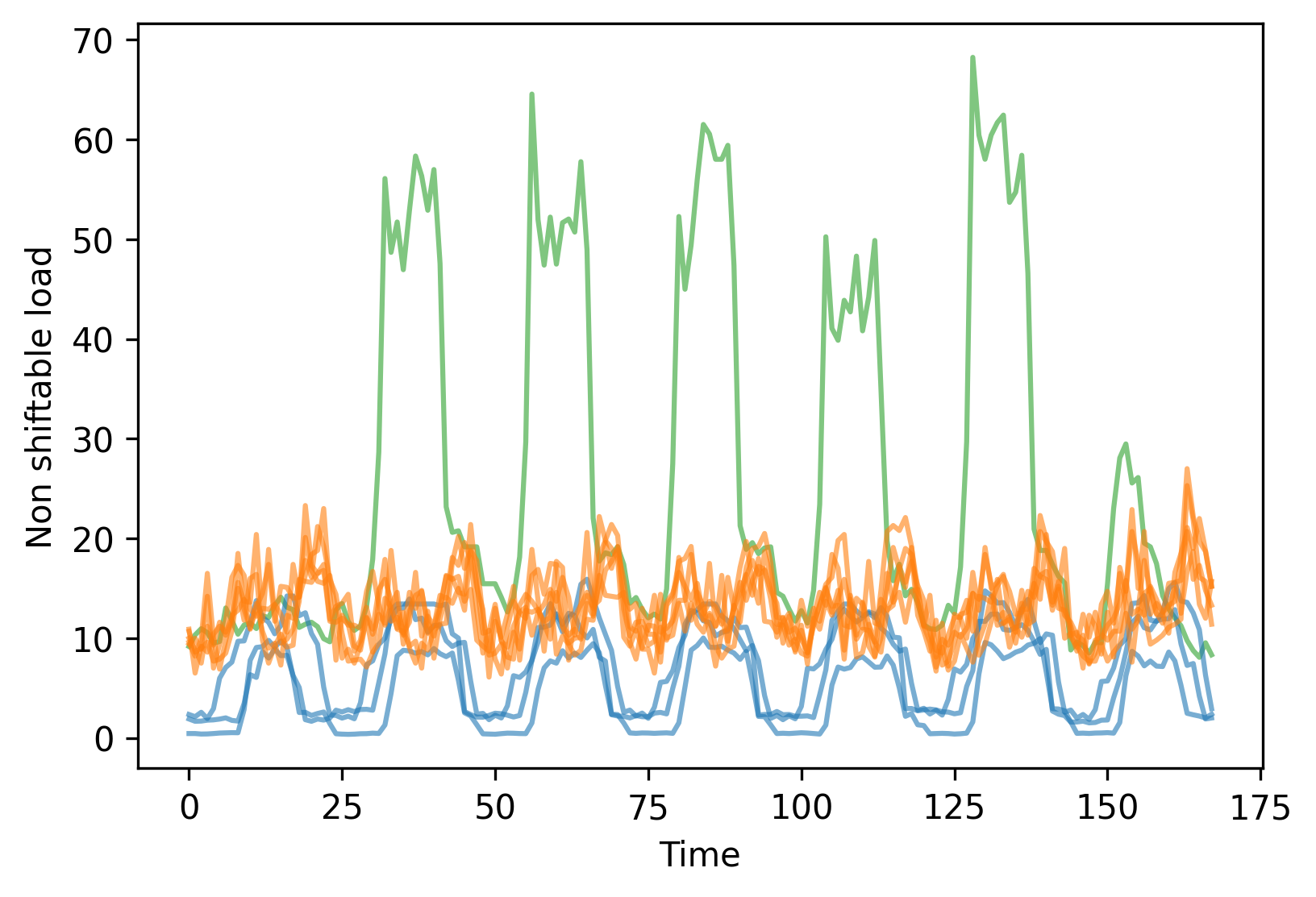}
    \caption{Behavioral clustering results.}
    \label{fig:clustering}
\end{figure}

% \begin{wrapfigure}{r}{0.5\textwidth}
%   \begin{center}
%     \includegraphics[scale=0.4]{figs/clustering1w.png}
%   \end{center}
%   \caption{Behavioral clustering results.}
%   \label{fig:clustering}
% \end{wrapfigure}

We retain 632 buildings, segmented into weekly episodes, forming independent RL tasks with shared structure but variable dynamics. Random task sampling during training increases meta-update frequency and prevents bias toward specific profiles. Figure~\ref{fig:clustering} visualizes clustering results on a subset of time series. To evaluate generalization, we hold out one behavioral cluster during meta-training and perform meta-testing on its buildings.
%This setup tests the model’s ability to adapt to previously unseen consumption patterns—assessing cross-distribution generalization within the same domain by leveraging structural regularities in building behavior.

\section{Experimental Evaluation}
\label{sec:experiments}

Experiments were conducted on two datasets: 1) a proprietary dataset of energy consumption data from 1,529 buildings collected between 2018 and 2024 ($\approx$ 30 million samples). It includes diverse building types (residential, office, industrial, and technical facilities) enriched with meteorological, socioeconomic, and electricity market data, and 2) the CityLearn open-source dataset \cite{vazquez2019citylearn}.

%\paragraph{Settings}
The goal in Building Energy Management Systems (BEMS) is to minimize energy costs and consumption by efficiently controlling energy flow. We use the publicly available CityLearn environment \cite{vazquez2019citylearn} as the simulation framework.
We consider a set of buildings, each equipped with a controller managing the energy stored in or released from its Energy Storage Unit (ESU). At each time step, the building’s demand is met through renewable generation $E_t^{\text{pv}}$, storage exchange $E_t^{\text{ESU}}$, and grid supply $E_t^{r}$, where $E_t^{\text{ESU}} > 0$ indicates charging and $E_t^{\text{ESU}} < 0$ discharging. For each building, $L_t$ denotes the non-shiftable load, $E_t^{th}$ the thermal energy consumption, $H_t$ the normalized state of charge of the ESU, and $v$ its capacity. Total consumption is thus $E_t = L_t + E_t^{th} + E_t^{\text{ESU}} + E_t^{\text{pv}}$.

\textbf{States.} Each building’s state vector $s_t \in \mathbb{R}^{30}$ includes financial, climatic, and load-related features such as energy prices, temperature, and solar irradiation.

\textbf{Actions.} We use a discretized action space indicating the proportion of capacity to charge or discharge. Invalid actions are masked using prior system constraints. The valid range is $[0, \frac{v_d-H_t}{v_d}]$ if $E^{\text{pv}} \geq E_t$, otherwise $[-\frac{\max(E_t,H_t)}{v_d}, \frac{v_d-H_t}{v_d}]$.

\textbf{Rewards.} The agent minimizes energy costs and grid ramping fluctuations. With time window $h$ and weights $\alpha_1, \alpha_2$, the reward is:%\\
\begin{equation}
r_t = -\left( \alpha_1 C^g E_t^{r} + \alpha_2 \sum_{t'=t-h}^{t} |E^r_{t'} - E^r_{t'-1}| \right)
\end{equation}

\subsection{Implementation and Training}

We use the \acrfull{ppo} algorithm \cite{DBLP:journals/corr/SchulmanWDRK17} as the inner loop \acrshort{rl} agent. The agent interacts with the environment for 100k steps, with policy updates performed every $2,048$ steps (fewer than 50 updates per episode).
The outer loop consists of $N=600$ batch of 3 tasks. Meta-model parameters are optimized using the Adam optimizer %\cite{kingma2014adam}
and gradient $\nabla_{\phi} L(\phi) = (\phi - \theta_i^K)$. 
To prevent early task oversampling, the probability of revisiting previously seen tasks is set to zero during the first $10\%$ of meta-training steps ($t_\phi < 0.1H$). Afterward, the revisit probability $\eta_{t_\phi}$ follows a polynomial growth schedule:
\begin{equation}
    \eta_{t_\phi} = \eta_0 + (\eta_{\text{max}} - \eta_0) \cdot \left(\frac{t_\phi}{H}\right)^p,
\end{equation}
where $\eta_0$, $\eta_{\max}$, and $p$ denote the initial value, maximum probability, and polynomial degree. This encourages exploration early on and gradually increases revisits, allowing reuse of specialized actor knowledge for complex, long-horizon tasks.
The shared \acrshort{fe} is a three-layer MLP with 64 neurons per layer and ReLU activations. Both actor and critic networks are fully connected with 64 neurons; the actor uses Tanh activation and the critic approximates the value function $V(s)=g(z;\theta^Q)$.
During meta-testing, the PPO agent is initialized with meta-learned parameters $(\phi^\psi, \phi^Q)$. Task-specific actor weights from meta-training are not reused, the actor is randomly initialized and adapts via the shared feature extractor $\psi$, leveraging transferable representations for unseen tasks.

\subsection{Experiments and Results}

\begin{figure}[!ht]
    \centering
    \subfloat[Performance over the full training horizon (Citylearn dataset)\label{fig:res-sub0}]{{\includegraphics[scale=0.35]{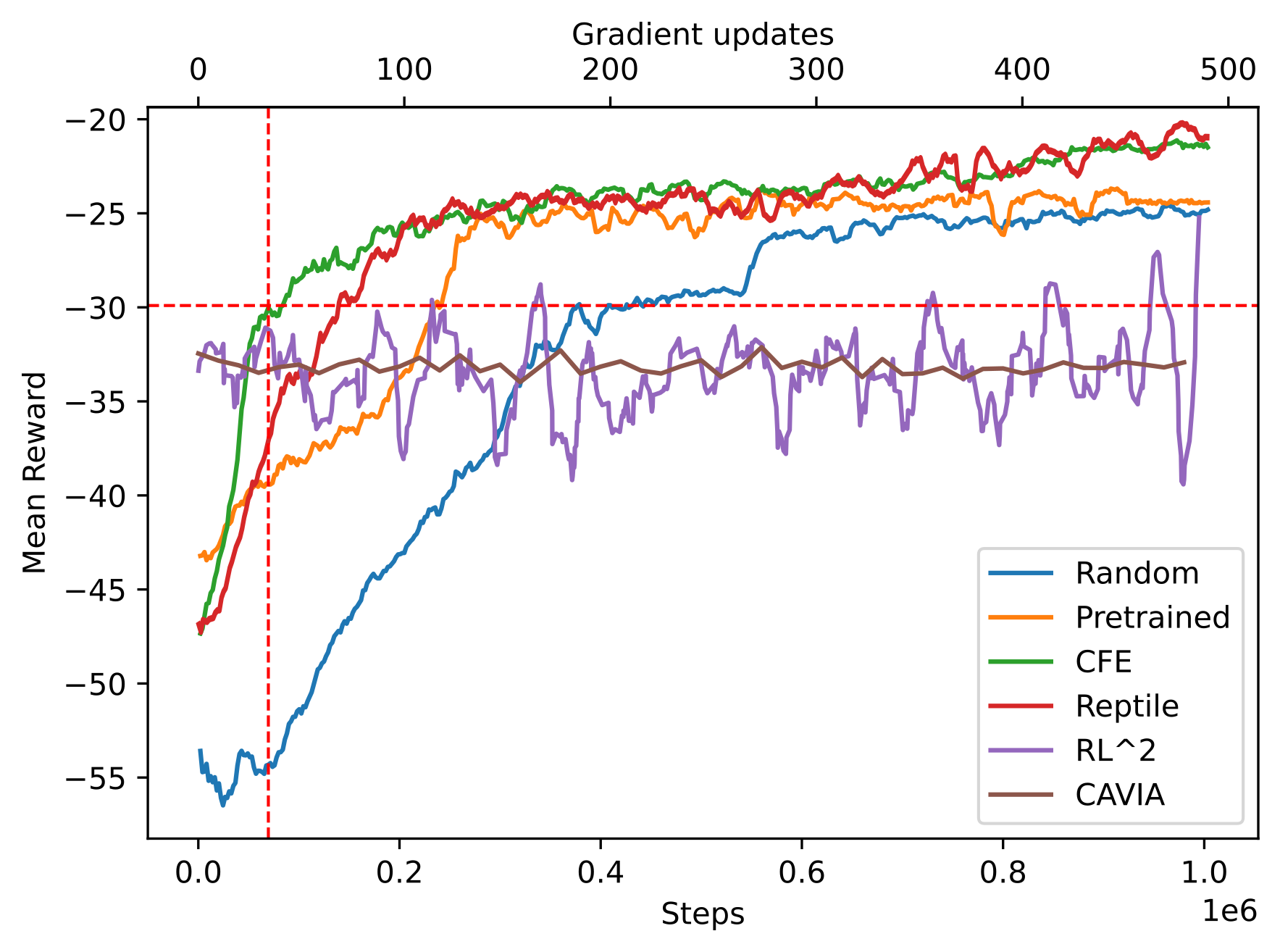}}}%
    \qquad
    \subfloat[Performance during early-stage adaptation (Citylearn dataset) \label{fig:res-sub1}]{{\includegraphics[scale=0.35]{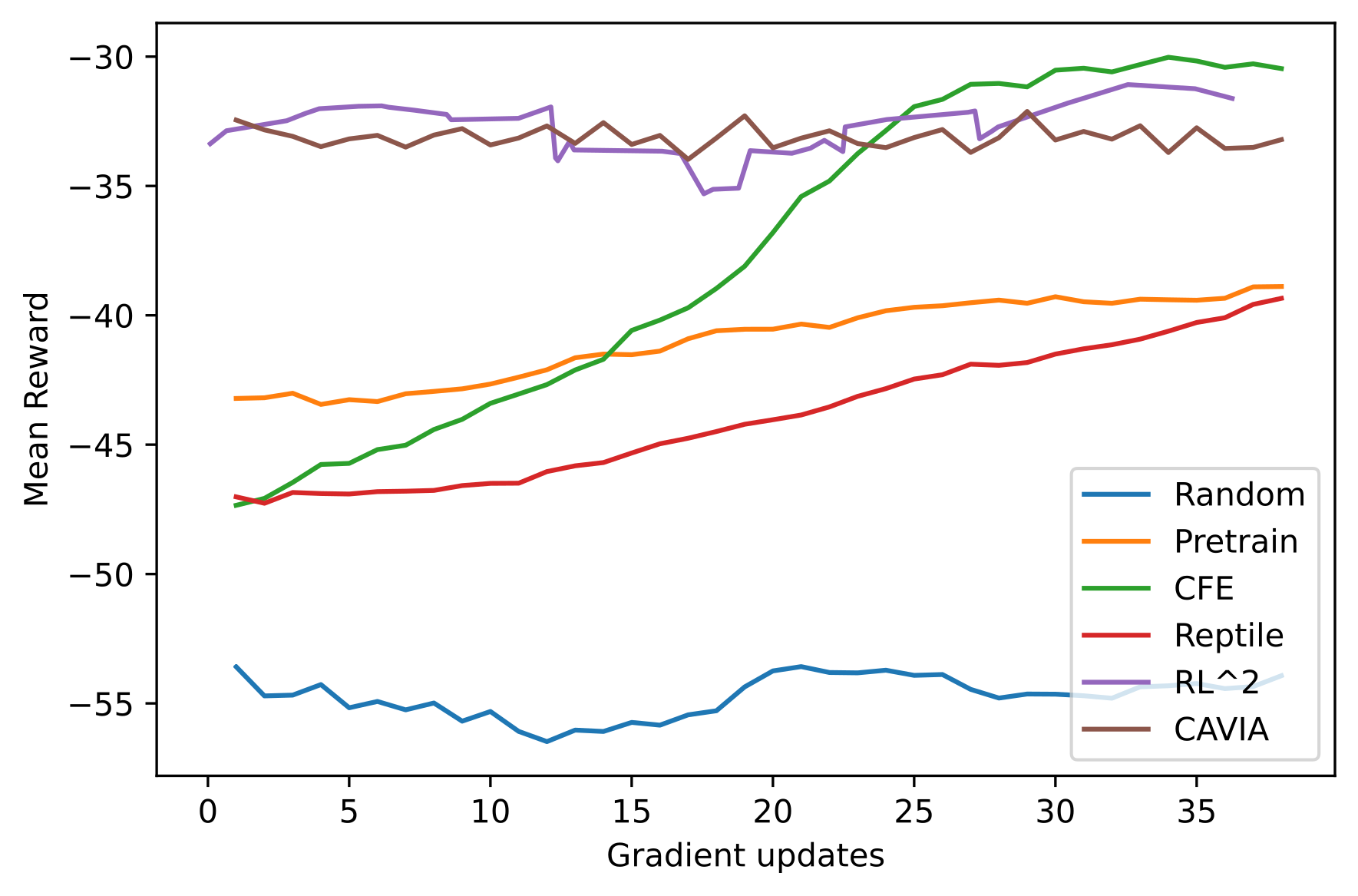}}}%
    \qquad
    \subfloat[\centering Performance on clusters at various distance from the training group (Proprietary dataset) \label{fig:res-sub2}
    ]{{\includegraphics[scale=0.40]{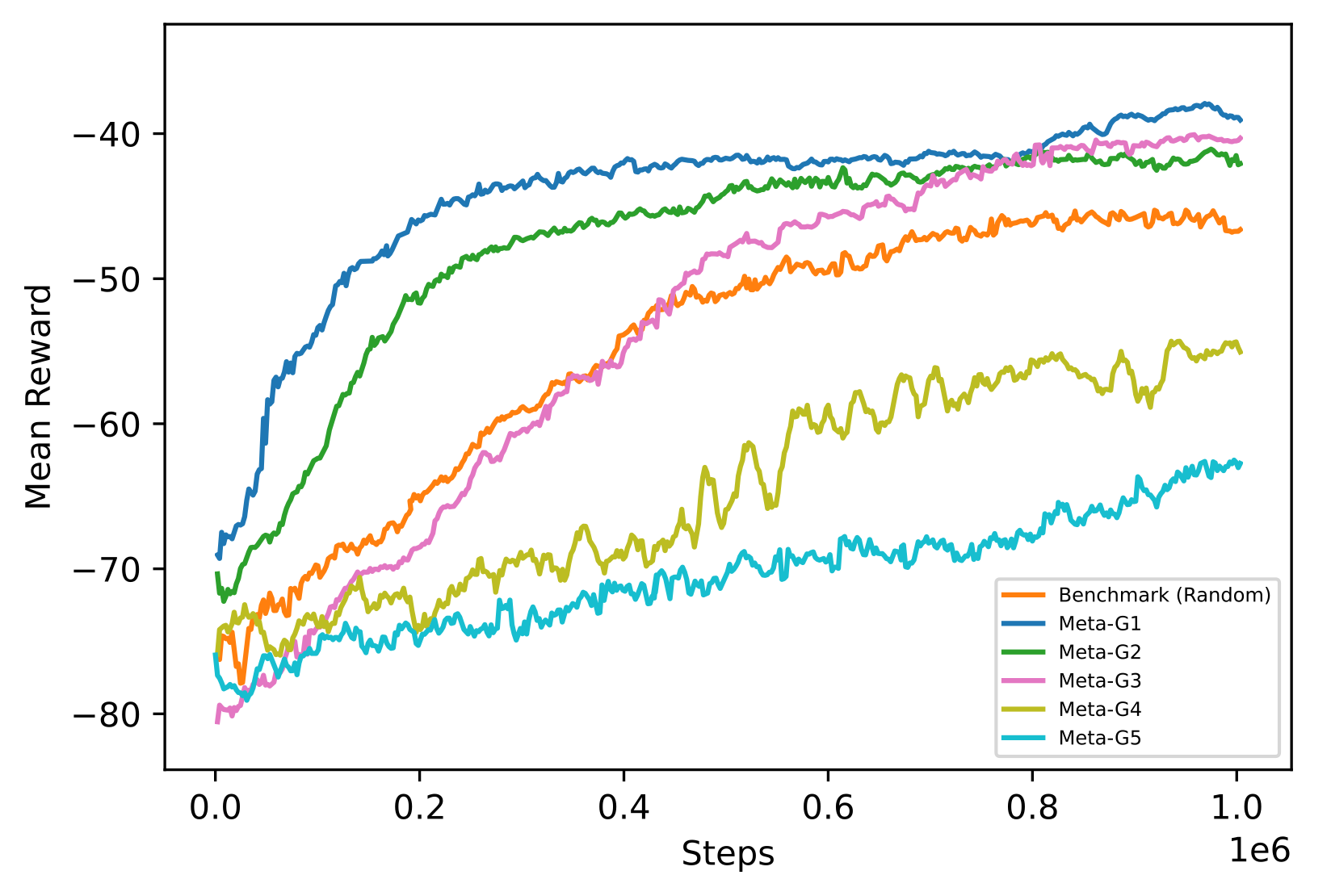} }}%
    \caption{Meta-testing initialization impact on unseen task (mean of 5 runs). The red dashed lines in (a) indicate the end of early meta gains shwon in (b).}%Red line represents the end of the meta-training gains.}%
    \label{fig:res}
\end{figure}

We compare our \textbf{CFE} approach against \textbf{CAVIA} \cite{zintgraf2019fast} and different PPO initializations: \textbf{Random} refers to a single-agent randomly initialized and trained from scratch until convergence on a specific building; \textbf{Pretrained} is a single-agent pretrained on one profile; \textbf{Reptile} refers to the vanilla Reptile without the \acrshort{fe} and actor reuse and \textbf{RL$^2$} uses a 256 layer LSTM to accumulate experience \cite{duan2016rl}.

\begin{figure}[!ht]
    \centering
    \subfloat[ Variance analysis \label{fig:std-meta}]{{\includegraphics[scale=0.3]{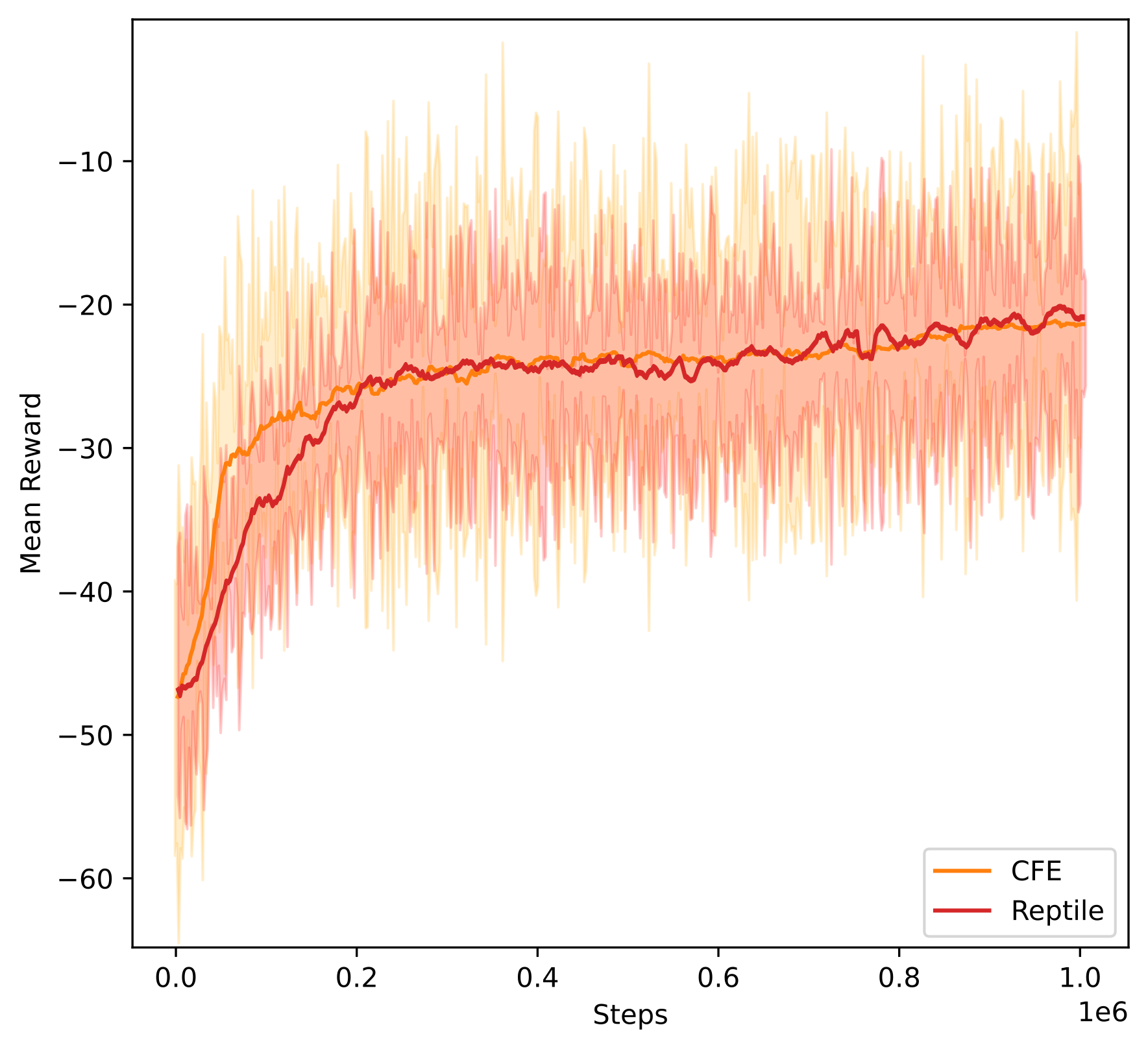}}}%
    \qquad
    \subfloat[\centering  Components ablation study  \label{fig:ablation}
    ]{{\includegraphics[scale=0.4]{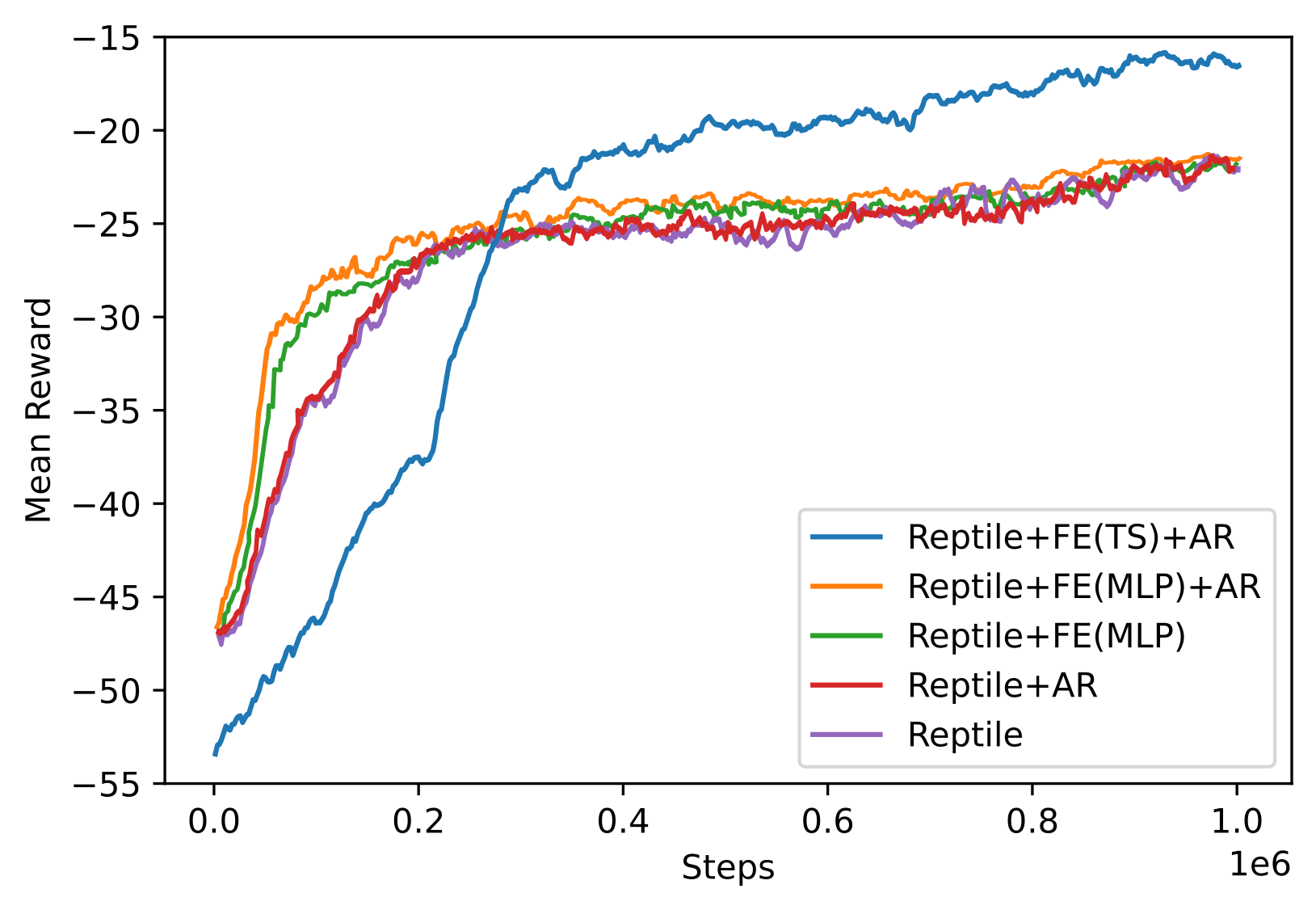} }}%
    \caption{Left: Variance of meta-trained agents across runs. Right: Ablation of Feature Extractors (FE: MLP or TS) and Actor Reuse (AR).}
\end{figure}

Figure~\ref{fig:res} presents the meta-testing results in terms of environment and gradient steps. Our Meta-RL agent reaches a mean reward of -30 in $\approx$70k steps, while the Pretrained and Random baselines require about 250k and 400k steps. The vanilla Reptile agent converges faster than both baselines but remains below our variant, confirming that our method accelerate convergence, though the slightly larger confidence interval (Fig.~\ref{fig:std-meta}) indicates reduced stability. Meta-trained agents benefit from parameters already aligned with the task distribution $p(\mathcal{M})$, requiring fewer updates to achieve strong performance on unseen tasks. In contrast, randomly initialized agents must explore extensively, and pretrained policies adapt more slowly to new dynamics. The CAVIA and RL$^2$ baseline shows stable performance but little improvement over episodes during meta-testing. They both behave as a robust generalist rather than a fast learner capturing shared structure but lacking effective within-task adaptation.\\
%This effect is clear in the early gradient-update analysis (Fig.~\ref{fig:res-sub1}). Our method and Reptile show rapid initial performance gains, while the Random and Pretrained baselines exhibit minimal improvement. Meta-initialization enables informed exploration early in training such as testing charge–discharge cycles, whereas Random agents still discover the state–action space, and Pretrained ones struggle to adjust their prior strategies. Overall, meta-learning provides a strong prior that reduces adaptation sample complexity by roughly a factor of four.
The early gradient-update analysis (Fig.~\ref{fig:res-sub1}) shows that our method and Reptile learn quickly in the first updates, unlike the Random and Pretrained baselines. Meta-initialization drives informed exploration, enabling rapid discovery of effective charge–discharge patterns, while the baselines struggle to adjust. This yields an adaptation sample complexity reduction of about four times.

Figure~\ref{fig:ablation} shows the contribution of the Actor Reuse (AR) mechanism and the Feature Extractor (FE) module.
We compare four variants: 1) vanilla Reptile, 2) Reptile+AR, 3) Reptile+FE, and 4) our full model (Reptile+AR+FE).
Results show that most performance gains come from the \acrshort{fe} module. Reptile and Reptile+AR achieve similar asymptotic performance, indicating limited impact from actor reuse alone. Adding the \acrshort{fe} consistently improves convergence speed and final rewards. Meta-learning transferable representations \textbf{in early layers} allows the agent to extract general features, benefiting both policy and value networks by emphasizing environment-level dynamics over task-specific patterns. 

To assess the impact of representation learning, we evaluate an alternative \acrshort{fe} built on a probabilistic transformer (TS). The transformer is jointly trained from actor–critic gradients and an auxiliary forecasting loss that predicts future values of a univariate consumption signal, providing a lightweight model-based inductive bias aligned with recent temporal world-model approaches \cite{zangato2024enhancing}. %This auxiliary task encourages the latent space to encode temporal regularities relevant for downstream control.
The TS-based Reptile variant achieves higher asymptotic performance, showing that richer temporal representations benefit long-horizon decision-making. However, adaptation becomes slower, likely due to the larger parameterization and increased depth between input embeddings and decision layers. Its adaptation performance matches the fully pretrained agent, showing that meta-learning still offers a beneficial initialization. Overall, this highlights a trade-off: transformer-based feature extractors can improve final performance but their size reduces the influence of the meta-update, reducing early adaptation speed.

\begin{figure}[!ht]
    \centering
    \subfloat[CFE (Reptile+AR+FE)\label{fig:gradients-ours}]{{\includegraphics[width=0.4\textwidth]{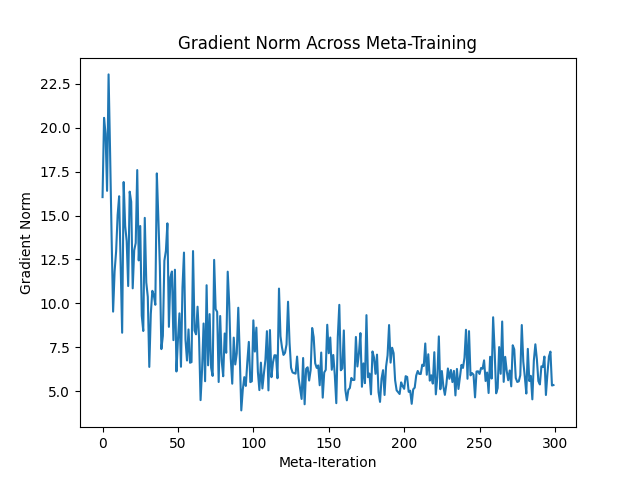}}}%
    \qquad
    \subfloat[\centering Reptile \label{fig:gradients-vanilla}
    ]{{\includegraphics[width=0.4\textwidth]{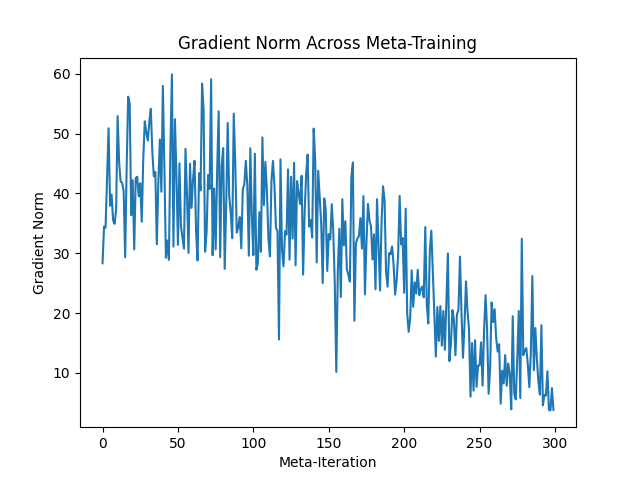} }}
    \caption{Evolution of the meta-gradient norm across training epochs for the standard Reptile algorithm and the proposed CFE variant.}
    \label{fig:gradient}
\end{figure}

Figure~\ref{fig:gradient} analyzes convergence by comparing the evolution of the meta-gradient norm between the Reptile and our CFE variant. Both exhibit high initial gradient norms, reflecting broad task exploration and rapid acquisition of general knowledge. Over time, the norm of CFE decreases faster, indicating quicker stabilization of meta-parameters. %Occasional spikes correspond to tasks with markedly different dynamics, momentarily increasing update magnitudes. 
This confirm that combining feature-level meta-RL with selective actor reuse accelerates convergence and enhances meta-stability, validating the efficiency of our proposed training framework.

We also evaluate the framework on our proprietary, more diverse dataset. Figure~\ref{fig:res-sub2} compares adaptation on building clusters at increasing distances from a reference trainign group (G0). Models trained on closely related clusters (Meta-G1, Meta-G2) achieve the highest performance, rapidly surpassing the random baseline and converging to stable policies with few updates. As cluster distance increases (Meta-G3), performance remains competitive but requires more training steps, indicating reduced transferability due to differing consumption patterns. For distant clusters (Meta-G4, Meta-G5), performance drops below the random baseline, suggesting that beyond a certain distance in the task distribution, shared knowledge becomes less relevant.
This shows the pattern-dependence of meta-learning: it provides strong benefits when source and target tasks share structural similarity but degrades as their underlying dynamics diverge.

To evaluate operational behavior, we analyze early-stage adaptation and performance metrics (Table~\ref{tab:cycles}). Our agent rapidly adopts strategic charge–discharge cycles, effectively exploring rewarding states and accelerating convergence. After only 15 gradient updates, it executes about five distinct cycles, rising to fifteen after 30 updates, while the randomly initialized agent performs roughly fifty unstructured cycles over the same period. This demonstrates the meta-policy’s capacity for meaningful exploration and adaptive control, in contrast to the rigid Pretrained agent, which exhibits limited flexibility.
We further assess performance using two operational metrics: (1) Ramping, defined as the absolute change in energy demand between consecutive timesteps, and (2) Financial cost, the electricity expense from grid usage. Results, normalized to a Rule-Based Controller baseline~\cite{vazquez2019citylearn}, show that our method consistently achieves lower ramping and energy costs than all baselines while matching Reptile’s performances.

\begin{table}[htbp]
\caption{Number of cycles of charge of the storing system and the final cost after $K$ rounds of policy updates.}
\label{tab:cycles}
\begin{center}
\begin{tabular}{|c|c|c|c|c|c|c|}
\hline
{$K$ Round} & \multicolumn{6}{|c|}{\textbf{Mean number of charging cycles}} \\
\cline{2-7} 
{of Updates} & {\textit{CFE (Ours)}} & Vanilla Reptile & RL$^2$ & CAVIA & {\textit{Random}} & {\textit{Pretrain}}\\
\hline
15  & $4.8 \pm 4.3$ & $6.2\pm 3.5$ & $17.5 \pm 5.6$ & $19.2 \pm 2$ & $58.6 \pm 12.4$ & $18.2 \pm 2.8$ \\
30  & $14.3 \pm 5.6$ & $14.8\pm 4.1$ & $17.3 \pm 4.2$ & $19.8 \pm 1.9$ & $45.3 \pm 15.3$ & $17.9 \pm 1.7$ \\
300 & $20.5 \pm 5.9$ & $18.9\pm 3.6$ & $18.1 \pm 5.4$ & $19.4 \pm 2.4$ & $16.4 \pm 4.6$ & $17.8 \pm 2.2$ \\
\hline
{} & \multicolumn{6}{|c|}{\textbf{Final Costs} ($^{\mathrm{a}}$Ramping, $^{\mathrm{b}}$Yearly financial cost)} \\
\hline
600 & \textbf{0.9}$^a$,\textbf{0.86}$^b$ & \textbf{0.9} $,0.87$ & 1.10,0.98 & 1.18, 1.02 & $1.01, 0.95$ & $1.03 \pm 0.96$ \\
\hline
%\multicolumn{5}{l}{$^{\mathrm{a}}$Ramping.$^{\mathrm{b}}$Yearly financial cost.}
\end{tabular}
\end{center}
\end{table}

%The faster learning and higher final performance achieved by the meta-testing agent demonstrate the sample efficiency of the proposed meta-learning approach. The meta-learned parameters serve as a robust prior that accelerates adaptation to new tasks, reducing the need for extensive task-specific exploration by retaining knowledge of the environment’s dynamics. Moreover, the observed performance gains highlight an effective balance between generalization and specialization: the framework leverages prior experience to generalize across diverse tasks while preserving the flexibility required to adapt to task-specific nuances.
\noindent Accelerated learning and improved final performance during meta-testing highlight the sample efficiency of our method. Meta-learning provide a strong prior that reduces task-specific exploration by encoding shared environment dynamics.

%%%%%%%%%%%%%%%%%%%%%%%%%%%%%%%%%%%%%%%%%%%%%%%%%%%%%%%%%%%%%%%%

\section{Conclusion}
We addressed key challenges in applying \acrshort{rl} to \acrshort{ems}, including sample efficiency, policy generalization, and adaptation to dynamic conditions. We proposed a Meta-RL framework for EMS control, integrating a first-order meta-learning algorithm into a hybrid actor–critic architecture. The method introduces two key innovations: 1) a shared feature extractor that promotes transferable representation learning across actor and critic networks, and 2) a task-specific actor reuse mechanism that accelerates adaptation to recurring EMS tasks.
Experiments demonstrate that our approach reduces adaptation sample complexity by a factor of four compared to standard RL, while achieving faster convergence and higher final performance than existing Meta-RL baselines. The model exhibits rapid early-stage adaptation without compromising long-term stability.
Nonetheless, the approach assumes structural similarity among tasks, which may limit generalization to out-of-distribution scenarios, and maintaining task-specific actor parameters introduces additional computational overhead. Future work will incorporate probabilistic latent task representations to enhance robustness and scalability under diverse conditions.

\bibliographystyle{splncs04}
\bibliography{biblio}
\end{document}